\documentclass{proc}
\usepackage[pdftex]{graphicx}
\usepackage{indentfirst}
\title{Hand Gesture Recognition with Leap Motion}
\author{
	Youchen Du,
	Shenglan Liu,
	Lin Feng,
	Menghui Chen,
	Jie Wu\\
	\texttt{Dalian University of Technology}}

\date{\today}
\begin{document}
\maketitle

\begin{abstract}
The recent introduction of depth cameras like Leap Motion Controller allows researchers to exploit the depth information to recognize hand gesture more robustly. This paper proposes a novel hand gesture recognition system with Leap Motion Controller. A series of features are extracted from Leap Motion tracking data, we feed these features along with HOG feature extracted from sensor images into a multi-class SVM classifier to recognize performed gesture, dimension reduction and feature weighted fusion are also discussed. Our results show that our model is much more accurate than previous work.

\end{abstract}

\section{INTRODUCTION}
In recent years, with the enormous development in the field of machine learning, problems such as understanding human voice, language, movement, posture become more and more popular, hand gesture recognition as one of the these fields has attracted many researchers's interest\cite{rautaray2015vision}. Hand is an important part of the human body, as a way to supplement the human language, gestures play an important role in daily life, in the fields of human-computer interaction, robotics, sign-language, how to recognize a hand gesture is one of the core issues\cite{ohn2014hand}\cite{wan2016hand}\cite{chaudhary2013intelligent}. In previous work, Orientation Histograms have been used to recognize hand gesture\cite{freeman1996computer}, a variant of EMD also have been used to finish this task\cite{ren2013robust}. Recently, a bunch of depth cameras such as Time-of-Flight cameras and Microsoft Kinect™have been marketed one after another, the use of depth features has been added to the gesture recognition based on low dimentional feature extraction\cite{suarez2012hand}. A volumetric shape descriptor have been used to achieved robust pose recognition in real time\cite{suryanarayan2010dynamic}, adding features like distance, elevation, curvature based on 3D information on the hand shape and finger posture contained in depth data have also improved accuracy\cite{dominio2014combining}. Recognize hand gesture through contour have also been explored\cite{yao2014contour}. Use finger segmentation to recognize hand gesture have been tested\cite{chen2014real}. Use HOG feature and SVM to recognize hand gesture have also been proposed\cite{feng2013static}.

The Leap Motion Controller is a consumer-oriented tool for gesture recognition and finger positioning developed by Leap Motion. Unlike products like the Microsoft Kinect™, it is based on binocular visual depth and provides data on fine-grained locations such as hands and knuckles. Due to the different design concepts, it can only work normally under close conditions, but it has a good performance on data accuracy with an accuracy of 0.2mm\cite{weichert2013analysis}. There have been many researches try to recognize hand gesture Leap Motion Controller\cite{ameur2017comprehensive}\cite{lu2016dynamic}. Combining Leap Motion and Kinect for hand gesture recognition have been proposed and achieved a good accuracy\cite{marin2014hand}.

Our contributions are as follows:
\begin{enumerate}
\item We propose a Leap Motion Controller hand gesture dataset, which contains 13 subjects and 10 gestures, each gesture by each subject is repeated 20 times, thus we have 2600 samples in total.
\item We use Leap Motion only. Based on the Leap Motion Controller part of [1], we propose a new feature called Fingertips Tip distance(T). With the use of the new feature, we observed considerable accuracy improvements on both of the data provided by [1] and our data.
\item We extract the HOG feature of Leap Motion Controller sensor images, HOG feature significantly improved gesture accuracy.
\item We explore different weight coefficients to combine the features of raw tracking data and the HOG feature of raw sensor images, by applying dimension reduction with principal component analysis, we finally get best accuracy of 99.62\% on our dataset.
\end{enumerate}

This paper is organized in this way: In Section II, we give a brief introduction of our model architecture, methods and our dataset. In Section III, we show the new Fingertips Tip Distance feature on the basis of previous work. In Section VI, we present the HOG feature extracted from sensor images from Leap Motion Controller. In Section V, we analyze and compare the performance of the new feature with the work presented by Margin et al, verifies the performance of the HOG feature alone, and evaluates the performance after applying PCA. In Section VI, we put forward the conclusion of this paper and thoughts on the following work.

\section{OVERVIEW}
In this section, we describe the model architecture we use and the way data is handled(Section 2.1), and how we collect our dataset by Leap Motion Controller(Section 2.2).

\subsection{SYSTEM ARCHITECTURE}

\begin{figure}
\includegraphics[width=\linewidth]{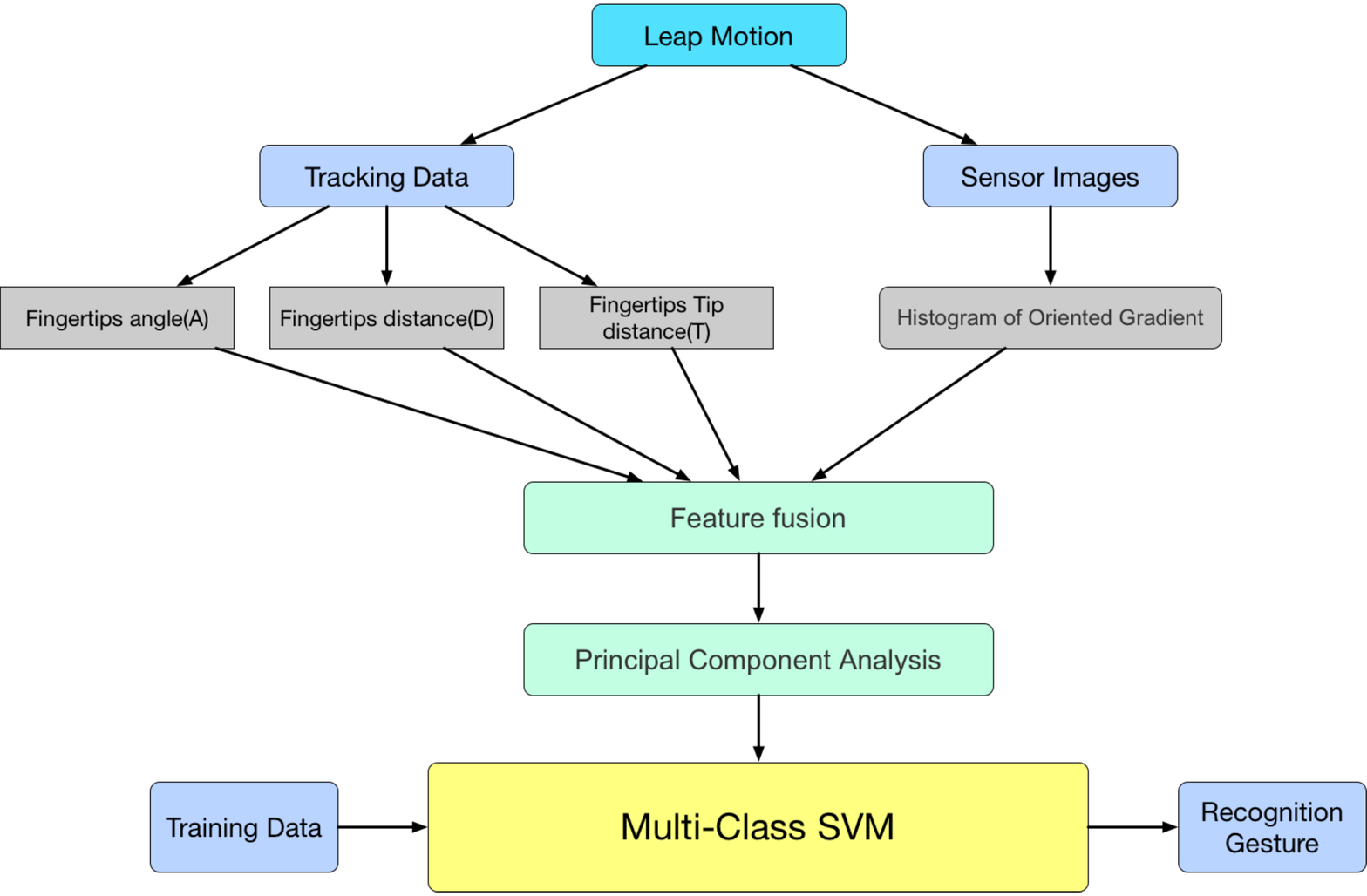}
\caption{System architecture}
\label{fig:system_architecture}
\end{figure}

Fig \ref{fig:system_architecture} shows in detail the recognition model we designed. The tracking data and sensor images of the gesture are captured simultaneously, for tracking data, we extract a series of related features, for sensor images, we extract the HOG feature, with some different weight coefficients, we apply feature fusion and dimension reduction by PCA, finally put these features into a One-vs-One multi-class SVM to classify hand gesture.

\subsection{HAND GESTURE DATASET}

\begin{figure}
\includegraphics[width=\linewidth]{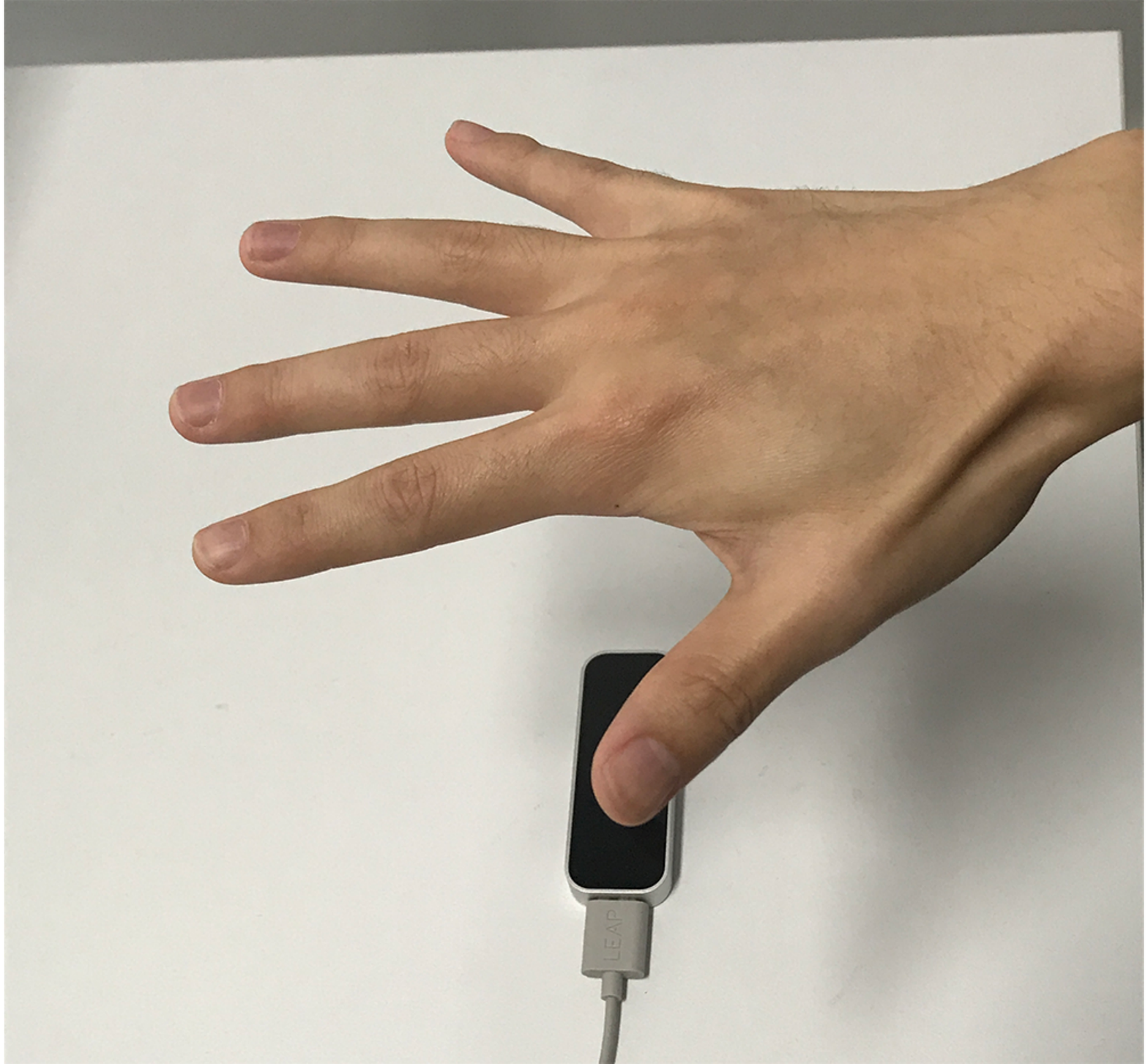}
\caption{Capture setup}
\label{fig:capture_setup}
\end{figure}

\begin{figure}
\includegraphics[width=\linewidth]{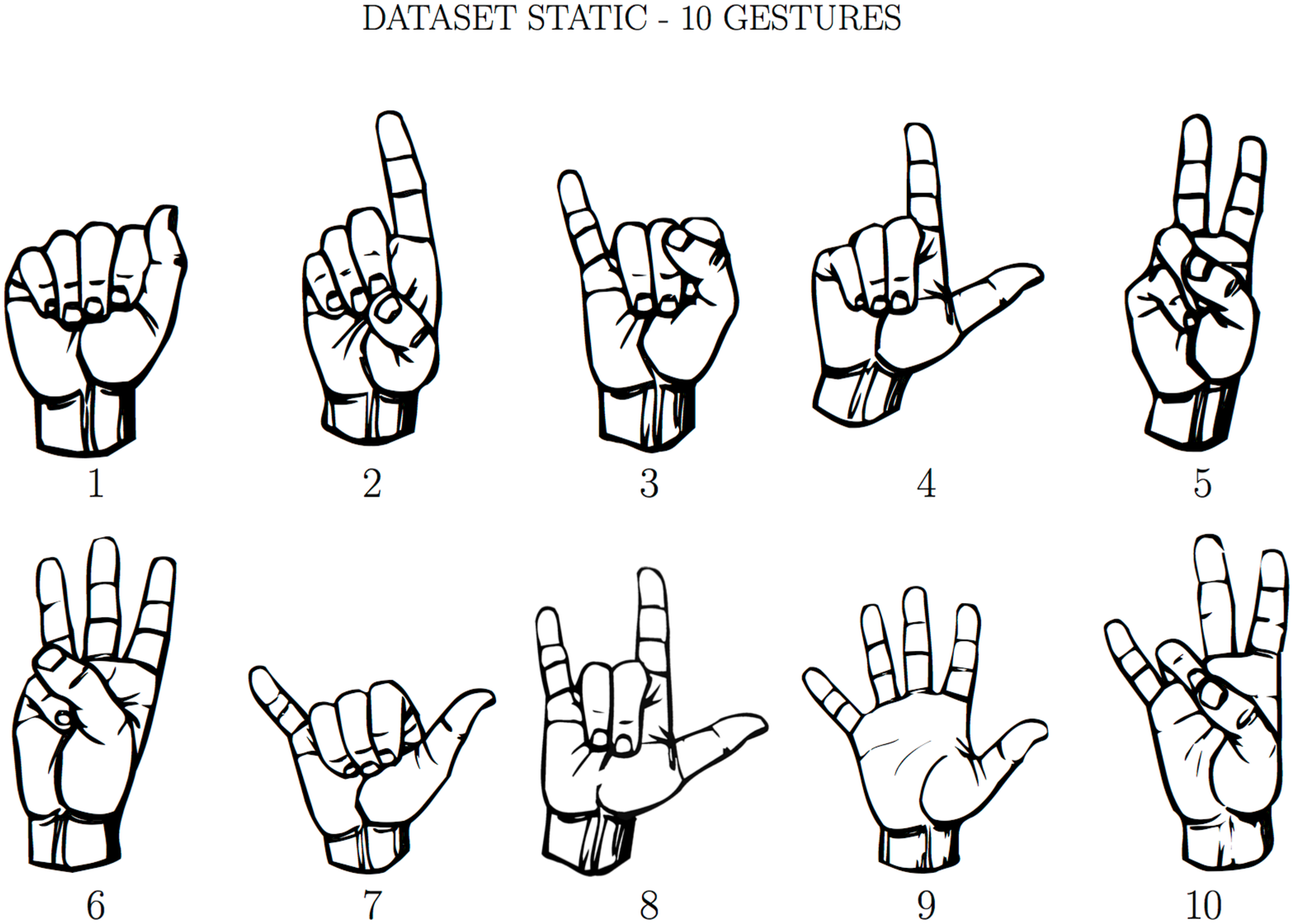}
\caption{Gestures in dataset}
\label{fig:gestures}
\end{figure}

In order to evaluate the effect of the new feature and test the HOG feature of the sensor images, we propose a new dataset, the setup is shown in Fig \ref{fig:capture_setup}. The dataset contains a total of 10 gestures(Fig \ref{fig:gestures}) performed by 13 individuals, each gesture is repeated 20 times, so the dataset contains a total of 2600 samples. The tracking data and sensor images are captured simultaneously, and each individual is told to perform gestures within Leap Motion Controller's valid 
visual range, allowing translation and rotation, no other prior knowledges.

\section{FEATURE EXTRACTION FROM TRACKING DATA}
Margin et al. proposed features like Fingertips angle(A), Fingertips distance(D), Fingertips elevation(E) from tracking data of Leap Motion Controller, as defined below:

\begin{equation}
\label{equ:A}
A_i = \angle(F^\pi_i-C,h)
\end{equation}
\begin{equation}
\label{equ:D}
D_i = \frac{\left \| F_i-C \right \|}{S}
\end{equation}
\begin{equation}
\label{equ:E}
E_i = \frac{sgn((F_i-F^\pi_i) \cdot n)\left \| F_i-F^\pi_i \right \|}{S}
\end{equation}

\begin{figure}
\includegraphics[width=\linewidth]{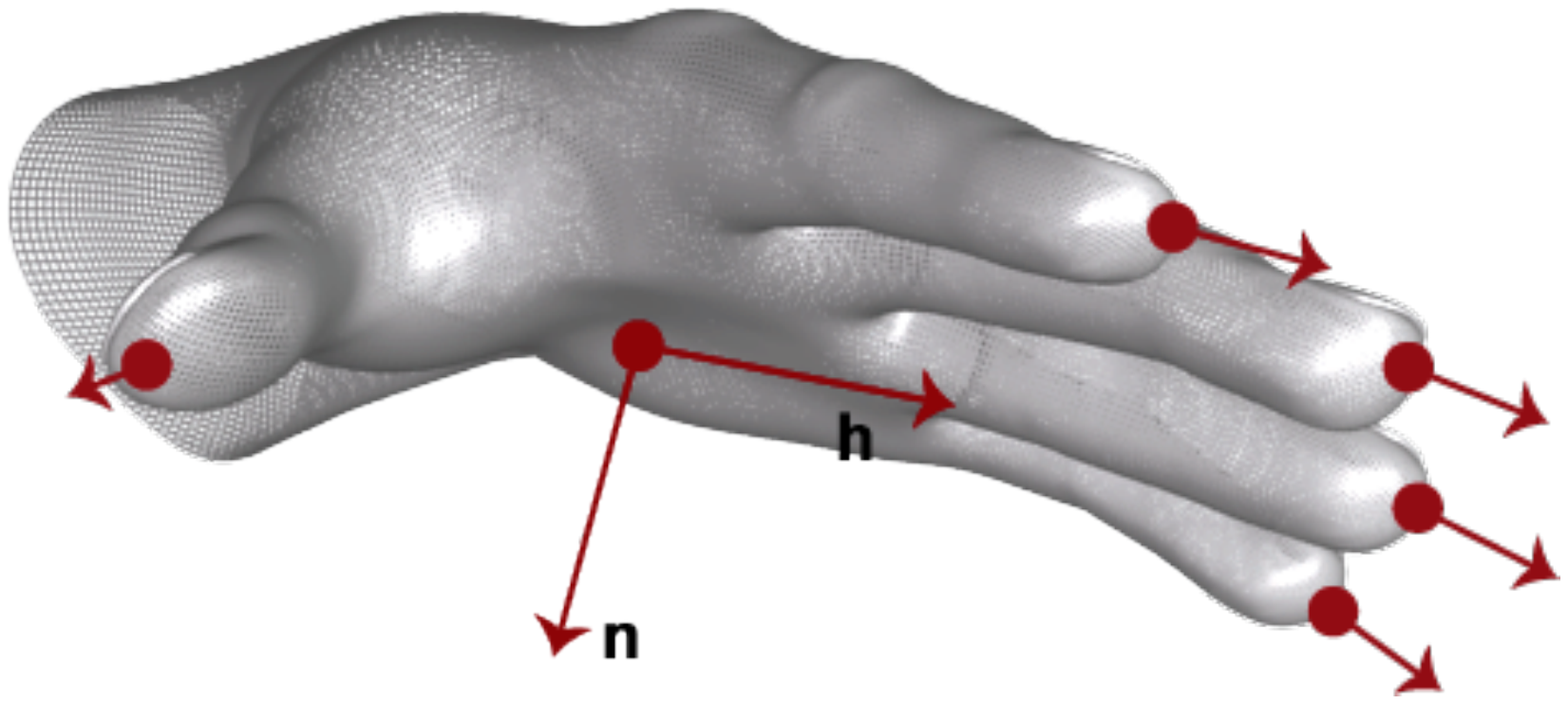}
\caption{Positions and Directions of fingertips}
\label{fig:hand_points}
\end{figure}

Where $S=F_{middle}-C$, $C$ is the coordinates of palm center in 3D space relative to the Controller, $h$ is the vector of directions from the palm center to the fingers, $n$ is the direction vector pointing perpendicular to the palm plane, as shown in Fig \ref{fig:hand_points} . $F_i$ is the coordinate of the detected fingertip, $i=1, \cdots, 5$, the data is not associated with any specified finger. $F^\pi_i$ is the projection of $F_i$ on the plane determined by $n$. $F_{middle}$ is the coordinates of the middle finger after associate $F$ with fingers. $A_i$ will be used to associate data to a specified finger, note that invalid $A_i$ will be set to 0 while the rest of the $A_i$ will be scaled to [0.5, 1], please refer to [1] for more details.

\begin{figure}
\includegraphics[width=\linewidth]{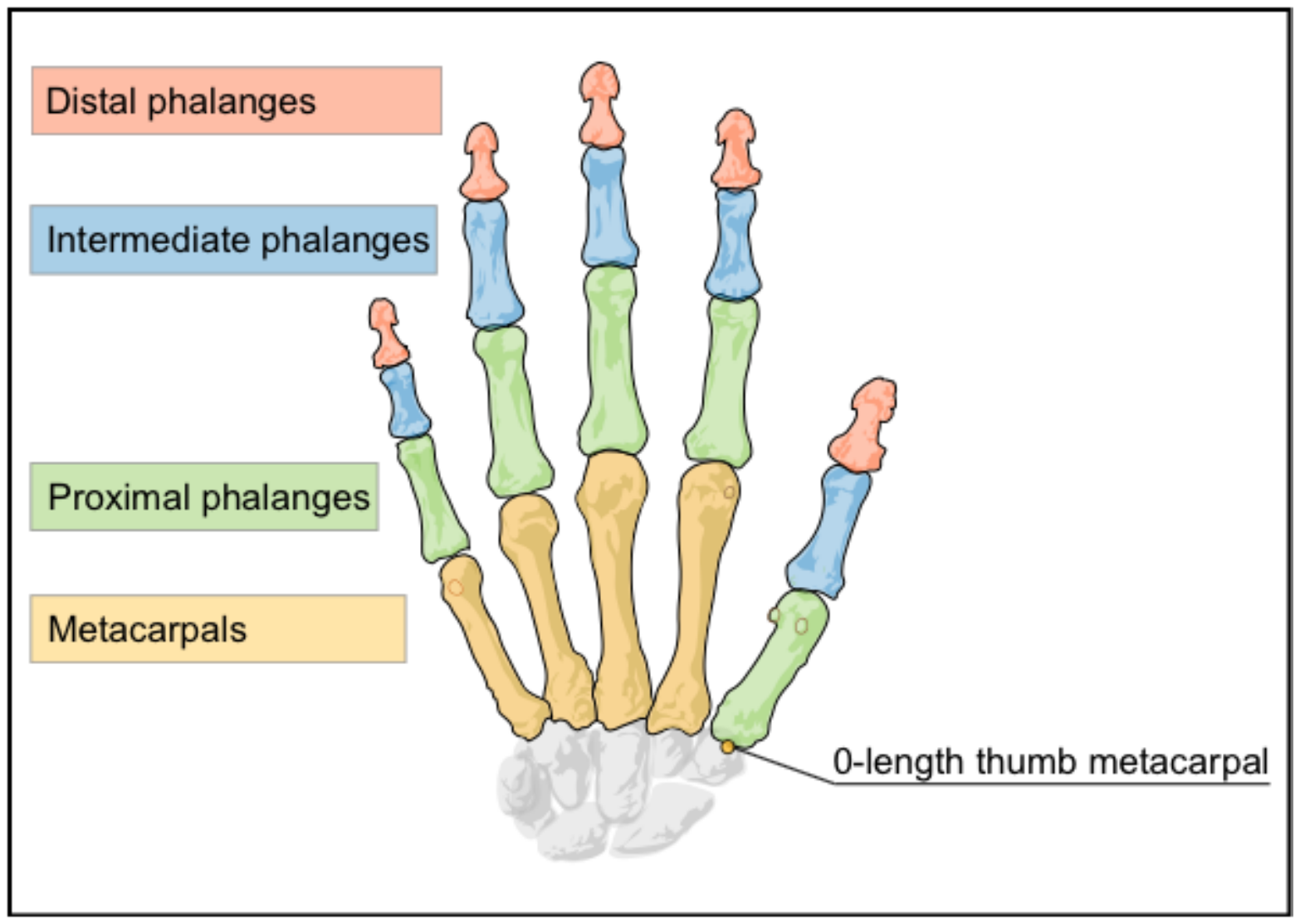}
\caption{The model of thumb from Leap Motion}
\label{fig:hand_joints}
\end{figure}

The previous study only considered the relationship between fingers and palm, and did not consider the possible relationship among fingers(Fig \ref{fig:hand_joints}). Therefore, we did some research for this part of work. By calculating the distance between fingertips and arranging them in ascending order of distance, we get a new feature we called Fingertips Tip Distance(T), as defined below:

\begin{equation}
\label{equ:T}
T_{ij} = \frac{\left \| F_i - F_j \right \|}{S}, 1 \leq i \neq j \leq 5
\end{equation}

After adding this feature, the performance has been further improved, the specific results will be given in Section V.

\section{FEATURE EXTRACTION FROM SENSOR IMAGES}
\subsection{Sensor images preprocessing}

\begin{figure}
\includegraphics[width=\linewidth]{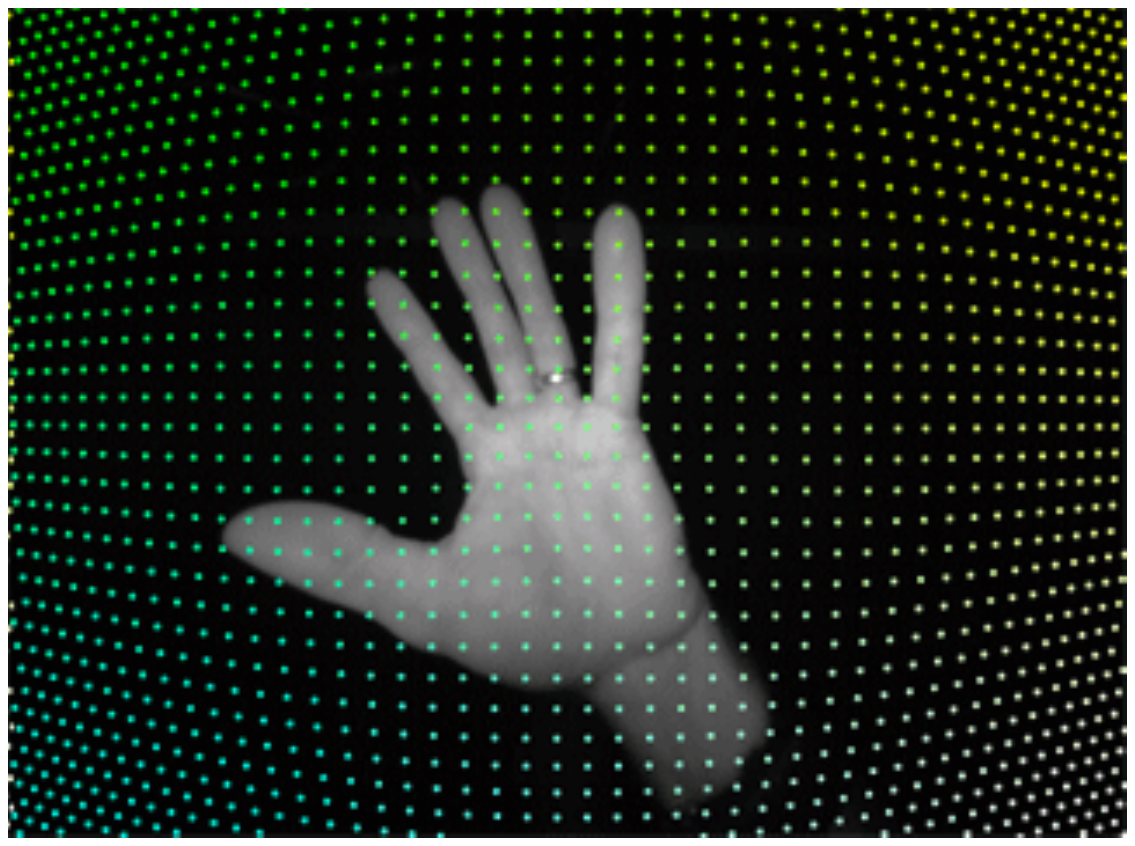}
\caption{Raw images from Leap Motion Controller}
\label{fig:leap_image_raw}
\end{figure}

Barrel distortion is introduced due to Leap Motion Controller's hardware(Fig \ref{fig:leap_image_raw}), in order to get realistic images we use a official method from Leap Motion to use bilinear interpolation to correct distorted images.

\begin{figure}
\includegraphics[width=\linewidth]{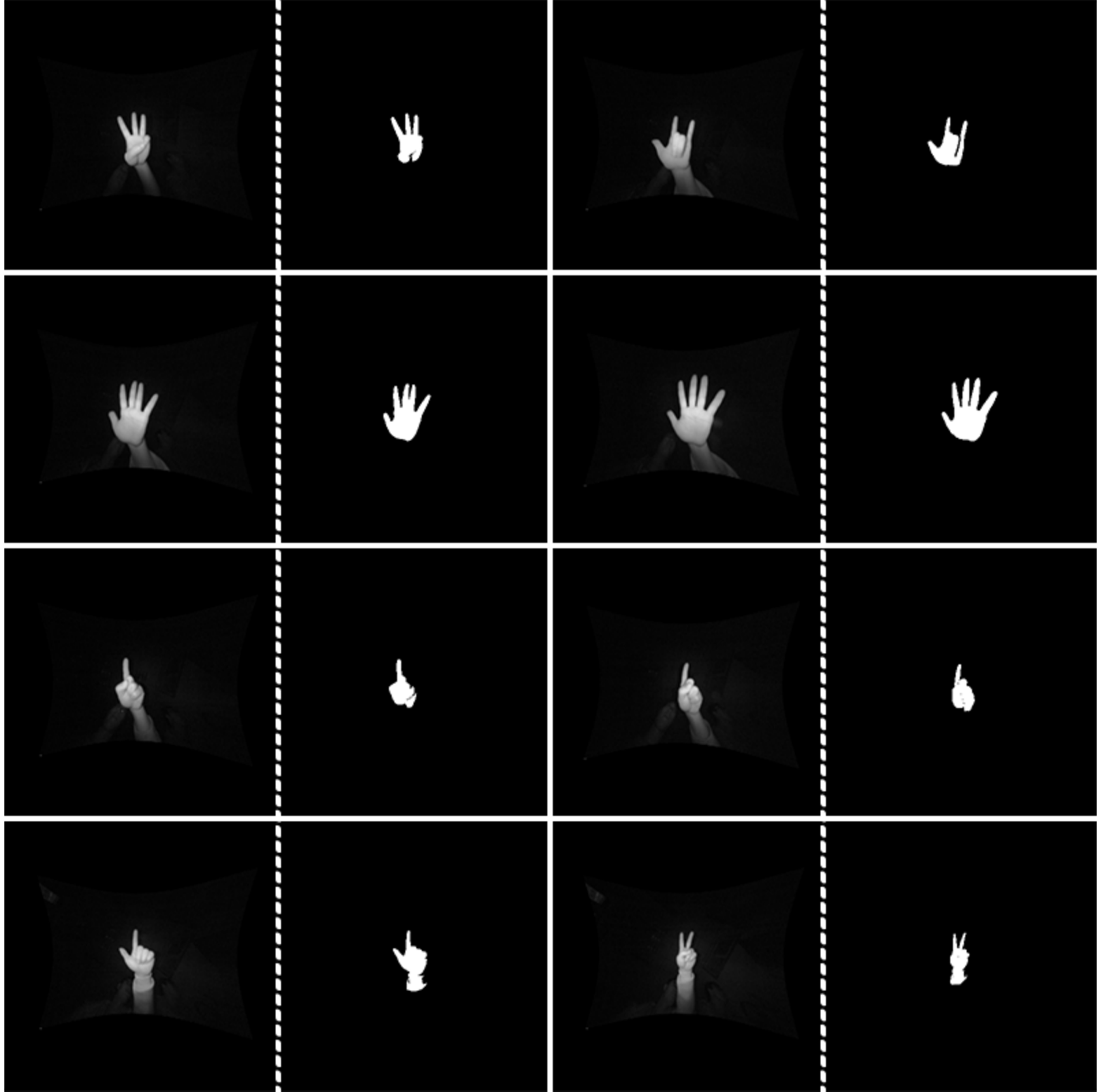}
\caption{Binarized images}
\label{fig:thres_compare}
\end{figure}

We use threshold filtering for the corrected image, and after doing so, the image will be binarized, retaining the area of the hand and removing the non-hand area as much as possible, as show in Fig \ref{fig:thres_compare}.

\subsection{Histogram of Oriented Gradient}

The HOG feature is a feature descriptor used for object detection in computer vision and image processing. Its essence is the statistics of image gradient information. In this paper, we use HOG feature to extract the feature information about gestures in binarized undistorted sensor images.

\subsection{Feature fusion}

For features like HOG from sensor images and ADET features from tracking data, we try to weight these features and feed into SVM to perform classification. we have tried to magnify the HOG feature differences by a factor of 1-9 and compare these results.

\section{EXPERIMENTS AND RESULTS}
\subsection{SVM architecture}

We use the One-vs-One strategy for multi-class RBF SVM to classify 10 classes, for each class pair there is a SVM, so result in a total of $10*(10-1)/2=45$ classifiers, the final classification result based on votes received. For hyper-parameters like $(C,\gamma)$, we use the grid search method on 80\% of the samples with 10-fold cross-validation, $C$ is searched from $10^0$ to $10^3$, $\gamma$ is searched from $10^{-4}$ to $10^0$.

For all the following experiments, each experiment was performed 50 times, and each sample was divided randomly according to the ratio of 80\% train set and 20\% test set, the final result is the average of the 50 experiments.

\subsection{Comparison among Tracking data features}
\begin{table}
\centering
\begin{tabular}{|c|c|c|}
\hline
Features Combination & Dataset from Marin et al. & Ours \\
\hline
D+E+T & 72.33\% & 59.50\%\\
A+E+T & 79.31\% & 83.37\%\\
A+D+E & 79.80\% & 82.30\%\\
A+D+T & 80.81\% & 83.60\%\\
A+D+E+T & 81.08\% & 83.36\%\\
\hline
\end{tabular}
\caption{Different features combination performance on different datasets}
\label{tab:tracking_data_features_compare}
\end{table}

We reconstruct the calculations of features like A, D, E based on equation \ref{equ:A}, equation \ref{equ:D} and equation \ref{equ:E}. 
We add the calculation for feature T based on equation \ref{equ:T}. Then we use the dataset from Marin et al. and our dataset to validate the Fingertips Tip Distance(T) feature, the results as shown in table \ref{tab:tracking_data_features_compare}.

\begin{figure}
\includegraphics[width=\linewidth]{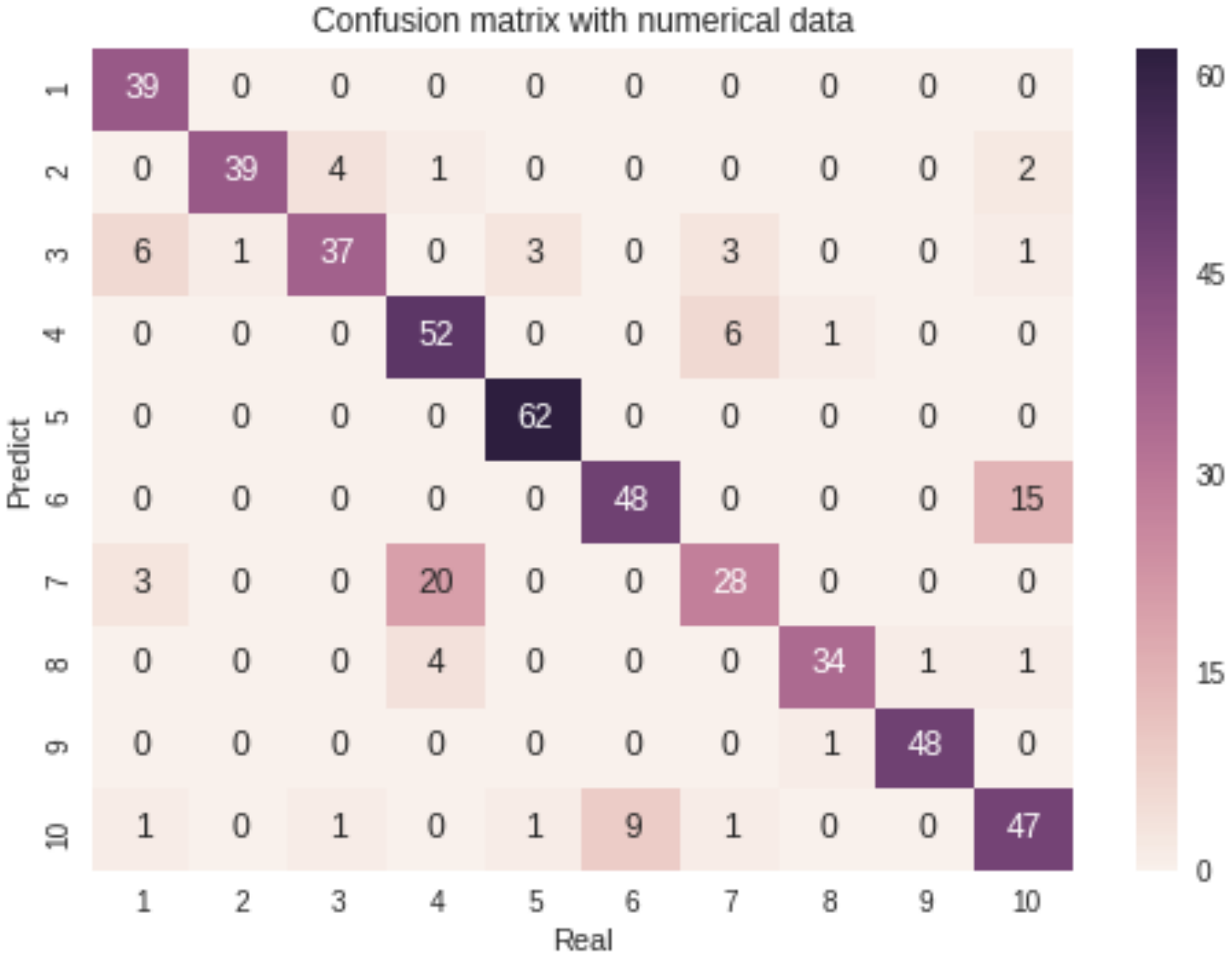}
\caption{A+D+T confusion matrix}
\label{fig:adt_mat}
\end{figure}

From table \ref{tab:tracking_data_features_compare}, we observe that both A+D+T and A+D+E+T outperforms A+D+E more than 1\% on both of datasets, whereas the difference between A+D+T and A+D+E+T is about 0.2\%, the difference is within normal range considering the different distribution of dataset. So it can be concluded that the feature combination of A+D+T is better than A+D+E, the confusion matrix from A+D+T is shown in Fig \ref{fig:adt_mat}.

\subsection{Feature fusion}

\begin{table}
\centering
\begin{tabular}{|c|c|c|c|}
\hline
 & A+D+T & HOG & A+D+T+1x HOG \\
 \hline
 Accuracy & 81.54\% & 96.15\% & 98.08\%\\
 \hline
 \end{tabular}
 \caption{Feature fusion}
 \label{tab:feature_fusion_compare}
 \end{table}
 
 We first try simply concatenate feature A+D+T from tracking data and HOG feature from sensor images(we call 1x HOG), and compare performance on our dataset, the results as shown in table \ref{tab:feature_fusion_compare}.

\begin{figure}
\includegraphics[width=\linewidth]{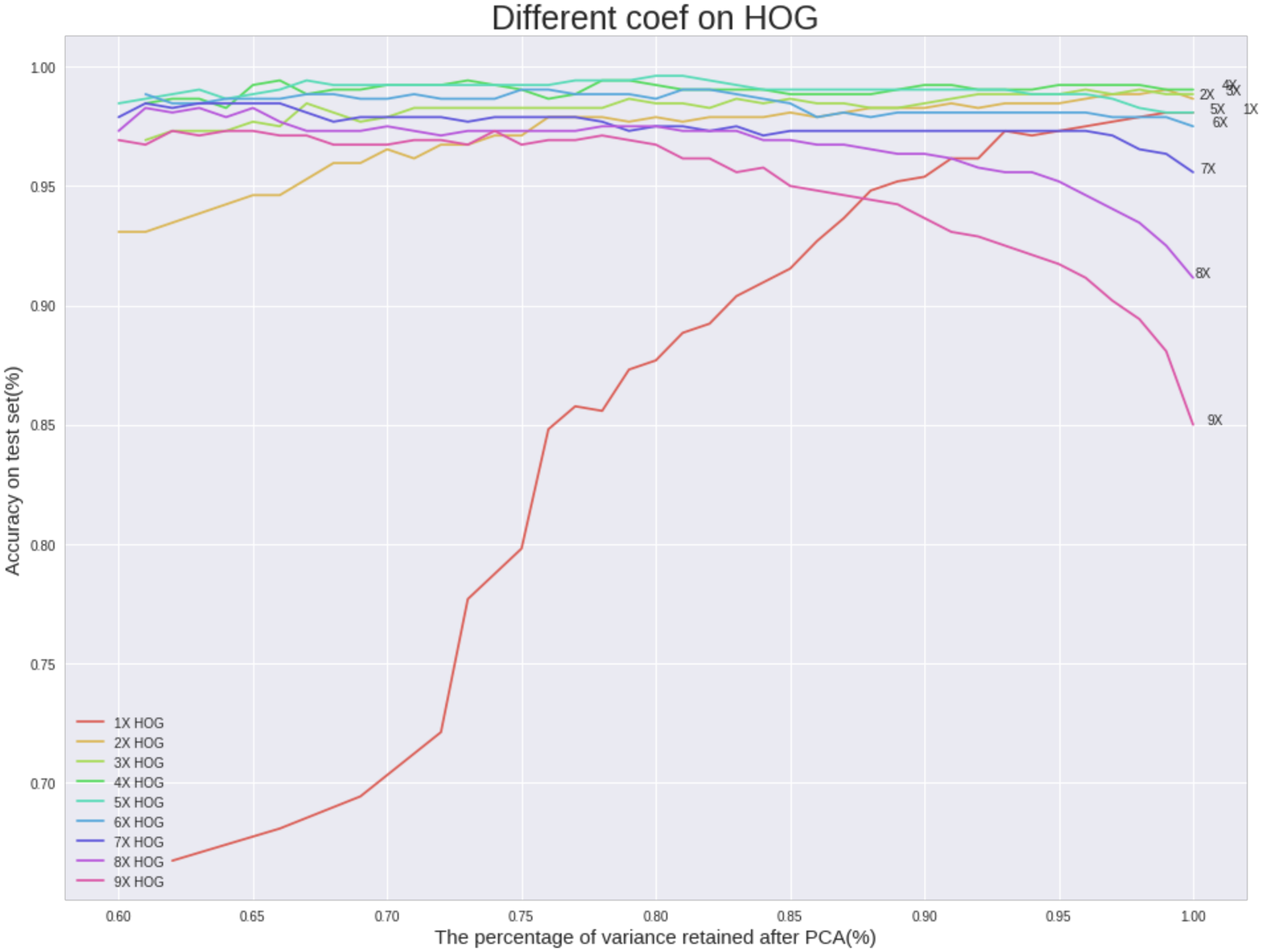}
\caption{Different weight coefficient for HOG feature}
\label{fig:different_coefs_on_hog}
\end{figure}

From table \ref{tab:feature_fusion_compare}, A+D+T+1x HOG outperforms single HOG about 2\%, we further explore the performance of different A+D+T+Kx HOG, $K=1, \cdots, 9$, the results as shown in Fig \ref{fig:different_coefs_on_hog}.

The experimental results indicate that the model achieves the best overall performance with 4x HOG and 5x HOG and retains the accuracy of about 98\% even when dimension is reduced to 60\% by PCA. When $K < 4$, as dimension decreases, the overall accuracy of the model shows a downward trend, especially 1x HOG. When $K > 5$, as dimension decreases, the overall accuracy of the model shows an upward trend, as can been seen from 9x HOG. We conclude that when $K = 4 or K = 5$, the variance of the features can be preserved even for significant dimension reduction by PCA, in our experiment, we only explore $K=1, \cdots, 9$, how to get a better $K$ still remains as a problem.

\begin{figure}
\includegraphics[width=\linewidth]{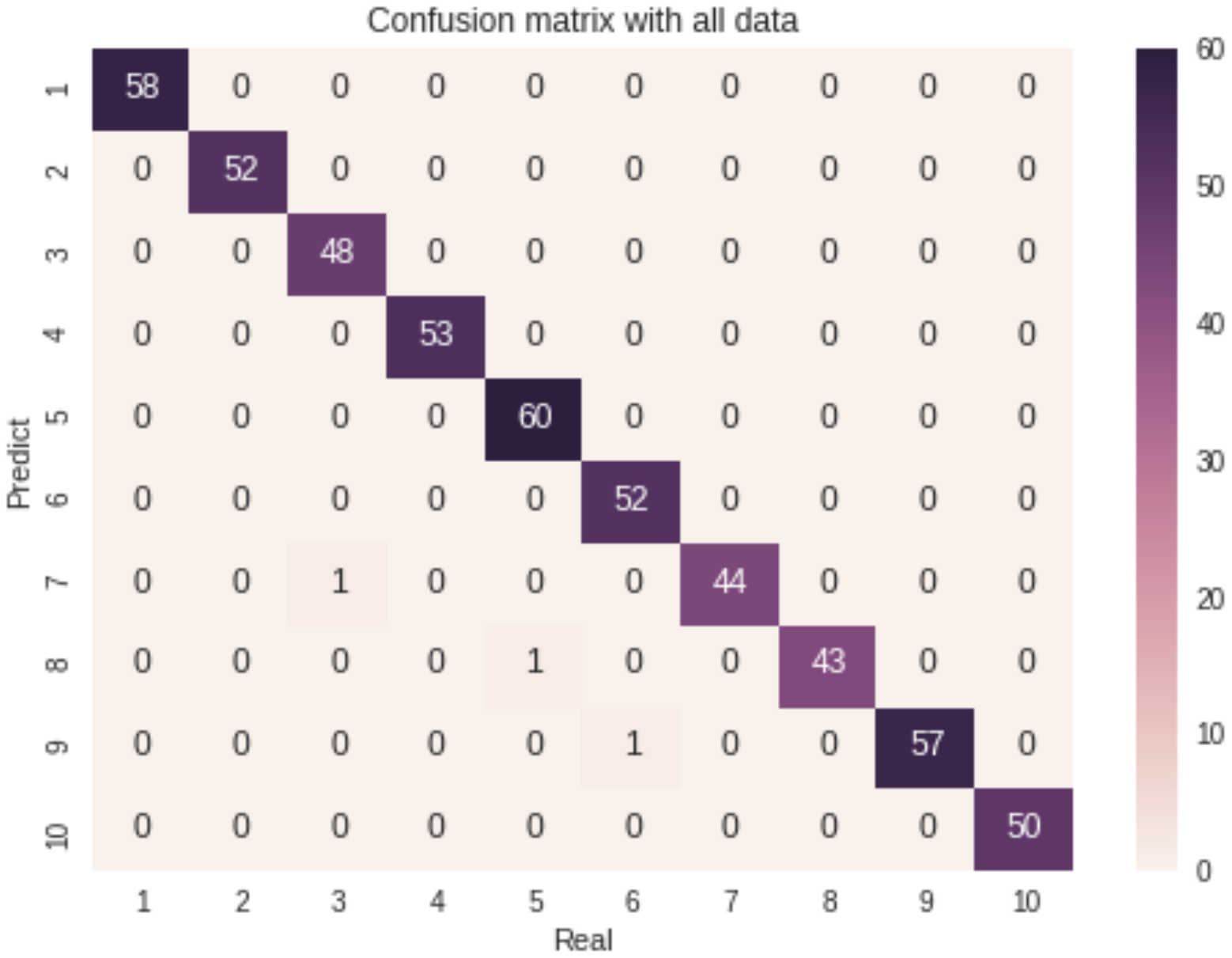}
\caption{Confusion matrix for A+D+T+5x HOG+80\% PCA}
\label{fig:best_fusion_confusion_matrix}
\end{figure}

Finally, we perform PCA preserving 80\% variance on A+D+T+5x HOG, and we get the best accuracy on our dataset of 99.42\%, the confusion matrix as shown in Fig \ref{fig:best_fusion_confusion_matrix}.

\section{CONCLUSIONS AND FUTURE WORKS}

In this paper, we deeply investigated the feature extraction of Leap Motion Controller tracking data, and propose a new feature called Fingertips Tip Distance which introduce 1\% more accuracy on two datasets. Meanwhile, we got higher accuracy by combining HOG feature extracted from binarized and undistorted sensor images and tracking data features. We also discussed the feasibility of linearly magnified HOG feature by different weight coefficients. Finally, we used PCA to perform dimension reduction on features with different coefficients.

In future work, we will further explore the characteristics of tracking data, we think the characteristics of the remaining joints will also affect the accuracy of the overall classification due to the correlation between joints. we perform feature fusion with different weight coefficients in this paper, the results are considerable, but how to get a better weight coefficient $K$ still remains as a problem which is also what we will do in the future. At the same time, we will study the interaction between the system and virtual reality application scenarios.

\bibliography{references}{}
\bibliographystyle{unsrt}
\end{document}